\pdfoutput=1
\documentclass[11pt]{article}

\usepackage[preprint]{acl}

\usepackage{amsmath}
\usepackage{amsfonts}
\usepackage{times}
\usepackage{latexsym}
\usepackage{amssymb}
\usepackage{algorithm}
\usepackage{algpseudocode}
\usepackage{balance}
\usepackage{makecell}
\usepackage[T1]{fontenc}
\usepackage[table]{xcolor}

\definecolor{lightgray}{gray}{0.9}
\definecolor{darkblue}{RGB}{94,110,186}
\definecolor{darkGreen}{RGB}{92, 148, 110}
\definecolor{darkgreen}{RGB}{0,100,0}
\definecolor{lightorange}{RGB}{255,160,0}
\definecolor{darkred}{RGB}{139,0,0}

\usepackage[utf8]{inputenc}

\usepackage{microtype}

\usepackage{inconsolata}

\usepackage{graphicx}
\usepackage{booktabs}
\usepackage{graphicx}
\usepackage{multirow}
\usepackage{url}
\usepackage{array}
\usepackage{stfloats}
\usepackage{comment}

\usepackage{xcolor}
\usepackage[normalem]{ulem}

\definecolor{lightgray}{gray}{0.9}
\definecolor{darkblue}{RGB}{94,110,186}
\definecolor{darkGreen}{RGB}{92, 148, 110}
\definecolor{darkgreen}{RGB}{0,100,0}
\definecolor{lightorange}{RGB}{255,160,0}
\definecolor{darkred}{RGB}{139,0,0}
\definecolor{OliveGreen}{rgb}{0.36, 0.71, 0.39}
\definecolor{cvprblue}{rgb}{0.21,0.49,0.74}

\title{\textsc{ReGATE}: Learning Faster and Better with Fewer Tokens in MLLMs}

\author{Chaoyu Li, Yogesh Kulkarni, Pooyan Fazli \\
         Arizona State University, Arizona, USA \\
         \texttt{\{chaoyuli, ykulka10, pooyan\}@asu.edu}\\[0.3em]
         {\normalsize \textcolor{cvprblue}{\url{https://people-robots.github.io/regate}}}}

\begin{document}

\maketitle

\begin{abstract}
The computational cost of training multimodal large language models (MLLMs) grows rapidly with the number of processed tokens. Existing efficiency methods mainly target inference via token reduction or merging, offering limited benefits during training. We introduce \textsc{ReGATE} (\textbf{Re}ference-\textbf{G}uided \textbf{A}daptive \textbf{T}oken \textbf{E}lision), an adaptive token pruning method for accelerating MLLM training. 
\textsc{ReGATE} adopts a teacher-student framework, in which a frozen teacher LLM provides per-token guidance losses that are fused with an exponential moving average of the student’s difficulty estimates.
This adaptive scoring mechanism dynamically selects informative tokens while skipping redundant ones in the forward pass, substantially reducing computation without altering the model architecture. Across three representative MLLMs, \textsc{ReGATE} matches the peak accuracy of standard training on MVBench up to \textbf{2$\times$ faster}, using only \textbf{38\%} of the tokens.\ With extended training, it even surpasses the baseline across multiple multimodal benchmarks, cutting total token usage by over \textbf{41\%}.
\end{abstract}

\section{Introduction}
\label{sec:intro}

Multimodal large language models (MLLMs) face significant challenges due to the high computational cost of training~\cite{jin2024efficient}.\ A key bottleneck is the self-attention mechanism, whose complexity grows quadratically with input sequence length~\citep{nips2017_attention}. This issue is especially severe in video tasks~\citep{videosavi, avatar}, where frames are tokenized into extremely long sequences.\ Consequently, training MLLMs on large-scale datasets requires substantial computing resources, limiting accessibility and slowing progress. Improving training efficiency is therefore essential for enabling MLLMs to scale to longer contexts, support richer multimodal inputs, and learn from larger and more diverse datasets that would otherwise be infeasible to process.

\begin{figure}[t]
    \centering
    \includegraphics[width=\linewidth]{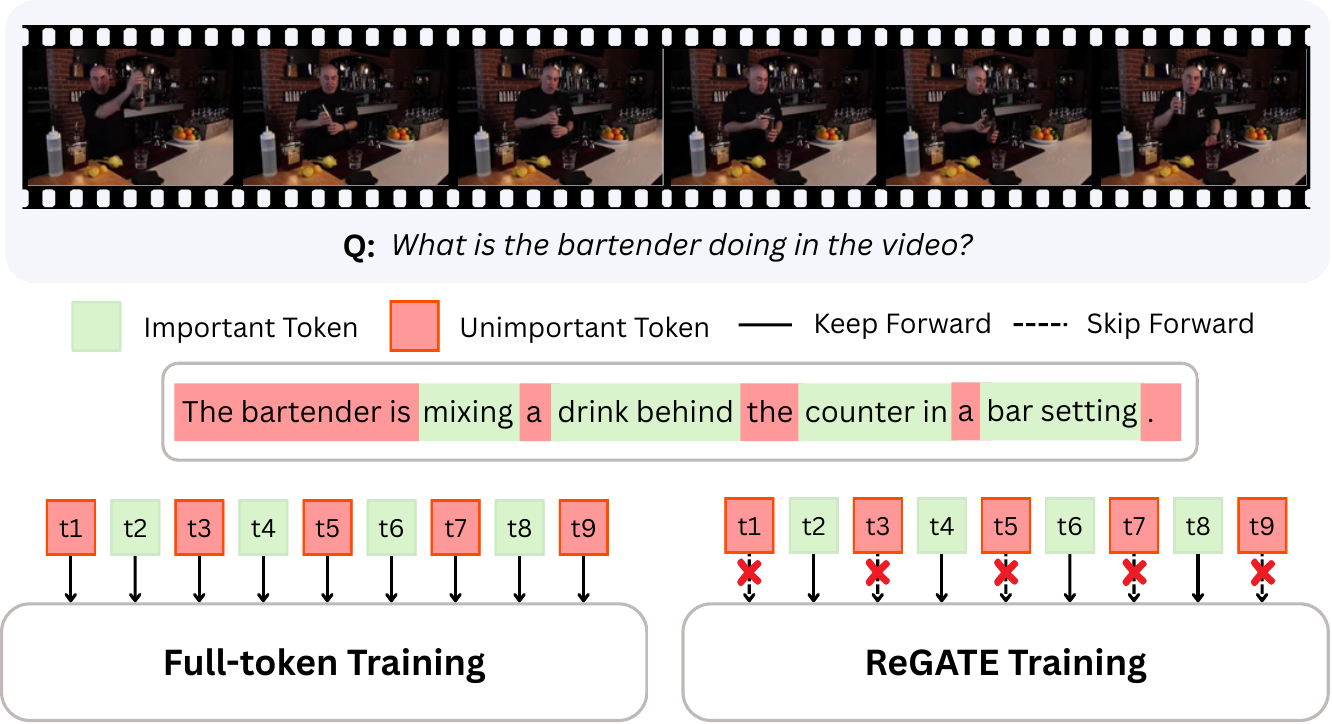}
    \caption{In training MLLMs, \textbf{\textsc{ReGATE}} identifies important textual tokens (light green) and selectively propagates them, while skipping unimportant ones (red).}
    \label{fig:teaser}
\end{figure}

Several strategies have been proposed to speed up inference in MLLMs, including token pruning~\citep{ye2024atpllava, arif2025hired, alvar2025divpruned, lee2025tamp}, token merging~\citep{chen2024fastv, dhouib2025pact}, and token compression~\citep{yang2025libra}. 
While these methods reduce FLOPs during inference, they rely on pre-scored or frozen tokens and cannot adapt token importance dynamically during training.
As a result, reducing training costs remains a more complex and less explored problem. 
In unimodal text models, learnable token pruning methods such as RHO-1~\citep{lin2024rho1} have improved training efficiency, but these approaches have not been extended to large multimodal models.
Early efforts to speed up visual processing, often focusing on standard vision transformers~\citep{akbari2021vatt} or early video-language models~\citep{lei2021clipbert}, rely on heuristics like random token dropping. In multimodal settings, however, determining which tokens are important often depends on subtle visual evidence,  temporal cues, or interactions between multiple modalities. Ignoring these dependencies can lead to the removal of crucial information, resulting in inefficient training and reduced multimodal understanding. This gap highlights the need for a principled method to estimate token importance during training, one that captures cross-modal dependencies without relying on heuristics.

To address this challenge, we introduce \textsc{\textbf{ReGATE}} (\textbf{Re}ference‑\textbf{G}uided \textbf{A}daptive \textbf{T}oken \textbf{E}lision), a framework designed to accelerate the training of MLLMs (Figure \ref{fig:teaser}). \textsc{ReGATE} adopts a teacher-student architecture, in which the student is the multimodal model being trained, and the teacher is a frozen, text-only version of the same LLM backbone. \textsc{ReGATE} combines two complementary signals to identify and retain the most informative tokens during training dynamically. First, it checks whether a token requires visual grounding by seeing if the text-only teacher can predict it from the prompt alone. Second, it evaluates the student model’s learning progress using an exponential moving average (EMA) of token-wise historical losses. By integrating these signals, \textsc{ReGATE} allocates computation to the subset of tokens that are both critical for multimodal understanding and remain challenging for the model to learn. To validate the effectiveness and generality of \textsc{ReGATE}, we evaluate it on three representative MLLMs: VideoChat2~\cite{li2024mvbench}, VideoLLaMA2~\cite{cheng2024videollama2}, and InternVL3.5~\cite{wang2025internvl35}, covering a wide range of architectures, training paradigms, and scales, and assess performance across diverse image and video benchmarks.

To summarize, our contributions are threefold:
\begin{itemize}
    \item We introduce \textsc{ReGATE}, an adaptive token pruning method for accelerating MLLM training. \textsc{ReGATE} leverages a text-only reference teacher model and the student’s historical token difficulty to dynamically identify and retain visually essential tokens, without introducing any additional trainable parameters.
    \item We show that the model-agnostic \textsc{ReGATE} integrates seamlessly into existing MLLMs without architectural changes or additional trainable components and remains effective across different transformer backbones under both full-parameter and LoRA fine-tuning.
    \item Extensive experiments on five image benchmarks and eight video benchmarks demonstrate \textsc{ReGATE}'s broad applicability and efficiency.\ On the challenging MVBench benchmark, \textsc{ReGATE} reaches the baseline’s peak accuracy up to \textbf{2$\times$ faster} while using only \textbf{38\%} of the tokens on average. 
\end{itemize}

\section{Related Work}
\label{sec:related-work}
 
\subsection{Token Compression for Fast Inference}

Most existing work focuses on accelerating inference, not training. 
Inference-time sparsity methods show that many tokens can be removed or merged with minimal impact on accuracy~\citep{shi2025staticdynamic, jiang2025visa, yang2025topv, hyun2025multigranular}. In vision transformers, Dynamic Token Pruning~\citep{tang2023dtop} halts processing of easy tokens layer by layer, reducing FLOPs by 20–35\% without degrading performance. For video LLMs, DyCoke~\citep{tao2025dycoke} compresses spatial-temporal tokens during inference, achieving up to 2$\times$ speed-ups while keeping model weights frozen. Moving beyond pruning, Importance-Based Token Merging~\citep{wu2025imagebased} merges similar tokens rather than dropping them, maintaining performance on long-video benchmarks while delivering 1.5$\times$ faster inference. However, all these methods operate after training is complete. During training, the full token is still processed in every forward and backward pass, leaving the computational cost of training mainly unaddressed.

\subsection{Token Compression for Fast Training} 

Only a few studies have explored token compression during training, rather than just at inference. In text‑only language models, RHO‑1~\citep{lin2024rho1} ranks tokens with a reference model and backpropagates only through the most difficult subset, reducing pre‑training tokens by 50\% while improving accuracy. For MLLMs, LaVi~\citep{yue2025lavi} injects vision-conditioned deltas into layer norms to skip visual tokens but needs extra modulation layers. LLaVA‑Meteor~\citep{li2025llavameteror} introduces a flash‑fusion module and a dual‑expert scorer that prunes 75–95\% visual tokens during instruction tuning but adds extra parameters and targets only visual tokens.
In contrast, \textsc{ReGATE} combines a \textit{static} cross-modal reference loss from a text-only teacher with a \textit{dynamic} EMA-based student signal. Together, these signals produce an adaptive, parameter-free sparsity mechanism that gates textual tokens while preserving visual ones, without modifying the model architecture.

\subsection{Teacher‑Student Distillation for MLLMs} 
Most distillation approaches for MLLMs mainly focus on parameter compression~\citep{li2023unlock,cai2025llavakd, udandarao2025activedatacuratione}. A systematic study~\citep{xu2024llavadi} shows that jointly aligning tokens and logits helps a smaller student model inherit visual grounding from a larger teacher model.
Similarly, methods like DIME‑FM~\citep{sun2023dimefm} show how cross-modal features can be transferred even from unpaired data. More recently, MaskedKD~\citep{son2024rolemasking} reduces teacher FLOPs by up to 50\% by masking patch tokens based on student attention, but it still backpropagates through all student tokens. In contrast, \textsc{ReGATE} leverages the teacher’s per-token loss to decide which tokens the student should process during each forward and backward pass. This approach significantly reduces computation without altering parameters and enables model-agnostic sparsity for more efficient training.


\section{ReGATE}
\label{sec:method}

\begin{figure*}[t]
 \centering
 \includegraphics[width=\linewidth]{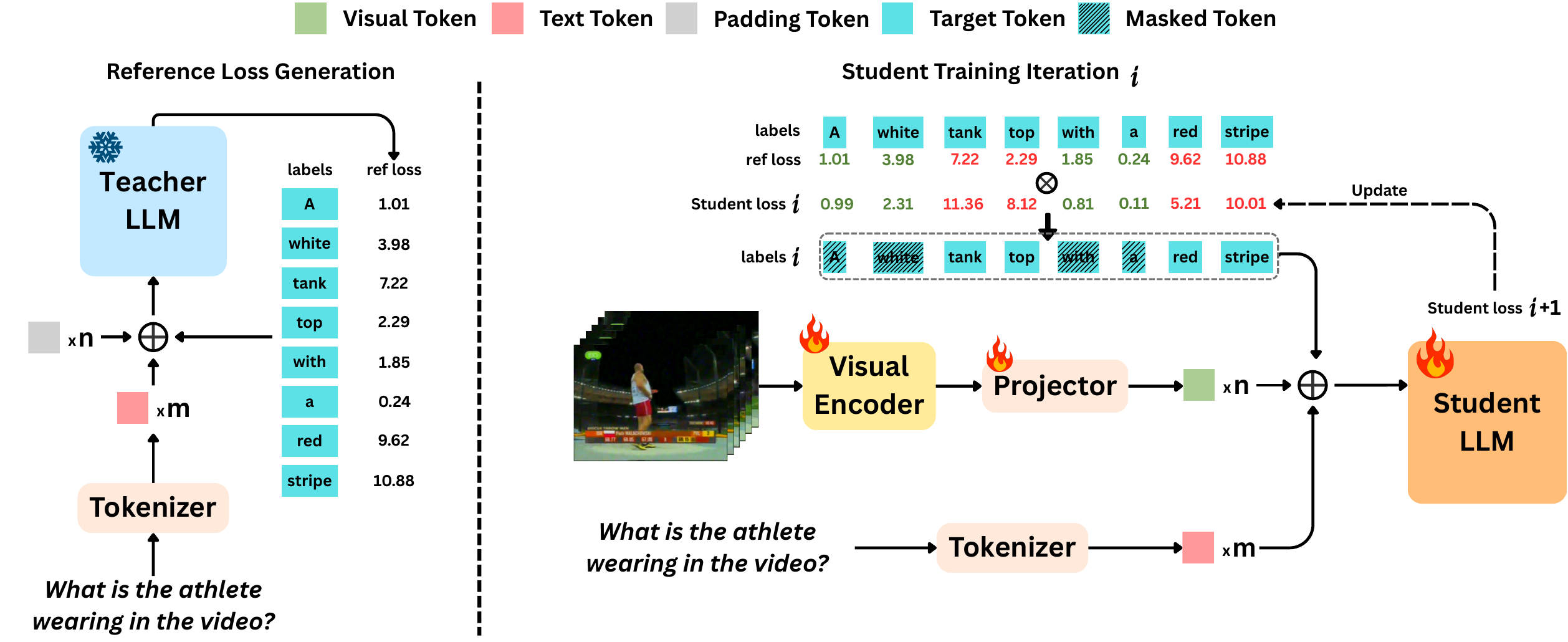}
 \caption{\textbf{Overview of \textsc{ReGATE}.} The framework operates in two interconnected stages. \textbf{(1) Reference Loss Generation (Left):} A frozen, text-only teacher LLM processes the input text (with padding tokens) and computes a per-token reference loss (\texttt{ref\_loss}), which measures how difficult each token is to predict from text alone. Higher loss values suggest the token likely requires visual grounding (e.g., ``white'', ``red stripe''). \textbf{(2) Student Training (Right):} The \texttt{ref\_loss} is combined with the student model’s historical learning difficulty to produce a unified importance score. This score is used to create a binary mask that selects the most informative tokens. During training, the student LLM receives the full multimodal input but only performs computation (e.g., self-attention and feed-forward operations) on the selected tokens, while skipping the rest. }
 \label{fig:method}
\end{figure*}

We introduce \textsc{ReGATE}, a method that accelerates MLLM training by allocating computational resources only to tokens that truly require visual information. The key insight is that not all tokens in a multimodal sequence depend equally on visual context: some can be accurately predicted from text alone, while others need cross-modal grounding.
To capture this, \textsc{ReGATE} uses a teacher-student framework. The student is the main MLLM being trained. The teacher is a reference model created by taking the student’s LLM backbone, removing its visual components (the visual encoder and projector), and freezing its weights. This results in a pure text-only LLM that acts as a fixed expert to estimate the degree to which each token depends on visual input.
Given a batch of input sequences containing both text and visual tokens, we generate a binary mask that determines which tokens should be actively computed and which can be skipped. This section explains how we calculate per-token difficulty scores using the frozen text-only teacher combined with the student’s own training history, how we dynamically adjust the fraction of tokens retained during training, and how we apply the resulting mask within the transformer decoder.

\subsection{Difficulty Score Formulation}
\label{sec:score-formulation}

Let $\mathbf{x}_b = (x_{b,1},\dots,x_{b,T})$ denote the token sequence in sample~$b$, including both text tokens and special visual tokens (e.g., \texttt{<image>} or \texttt{<video>} tokens representing visual content). To compute the reference loss, we construct a modified sequence $\hat{\mathbf{x}}_b$ by replacing the actual visual tokens with placeholder tokens (typically the padding token \texttt{<pad>}), ensuring the sequence length remains identical to the original multimodal input fed to the MLLM's backbone LLM. Our reference model is a pure text-only LLM obtained by removing the visual encoder and projector from the MLLM backbone, thus incapable of processing any visual content. By feeding the constructed placeholder sequence $\hat{\mathbf{x}}_b$ to the reference model in evaluation mode, we compute the per-token negative log-likelihood:

\begin{equation}
\ell^{\mathrm{ref}}_{b,i} = -\log p_{\mathrm{teacher}}\bigl(x_{b,i}\mid \hat{\mathbf{x}}_{b,<i}\bigr).
\end{equation}

A low value of $\ell^{\mathrm{ref}}_{b,i}$ indicates that the teacher can predict $x_{b,i}$ based on the textual context alone, whereas a high value signals that multimodal information is needed to predict the token.
In parallel, we monitor how difficult each token has been for the student across training updates. For every training sample $s$ and token position $i$, we maintain a running difficulty buffer $m_{s,i}$ updated as an exponential moving average (EMA) of the student's cross-entropy loss:
\begin{equation}
m_{s,i} \leftarrow \beta\, m_{s,i} + (1-\beta)\,\ell^{\mathrm{stu}}_{b,i},\quad \beta\in (0,1),
\end{equation}
where $\ell^{\mathrm{stu}}_{b,i}$ is the current cross-entropy loss of the student model at token position $i$, and $\beta$ controls the smoothing of the EMA. A higher value of $m_{s,i}$ indicates that token $i$ in sample $s$ has consistently posed difficulties during training.
We then combine the reference loss and the student's historical difficulty into a unified score for each token:
\begin{equation}
d_{b,i} = m_{s,i} + \lambda\,\ell^{\mathrm{ref}}_{b,i},
\label{eq:combined}
\end{equation}
where $\lambda$ balances these two signals. Tokens with a higher combined difficulty, $d_{b,i}$, either consistently challenge the student model or genuinely require visual context, and thus are prioritized during the training updates. 
Note that this combined difficulty evaluation is performed exclusively on output tokens (labels), as these tokens directly influence the training process through backpropagation.

\subsection{Dual‑cycle Sparsity Schedule}
\label{sec:dual-cycle}

We employ a deterministic schedule to determine the fraction of tokens kept at each training step. Our schedule repeats every $C$ steps. In the first $F$ steps of each cycle, we keep all tokens (i.e., $p=1$) to allow the model to stabilize. In the remaining $C-F$ steps, we retain only a fixed proportion $p_{\text{sparse}}$ of the tokens. Formally, if $t$ denotes the global training step, we have:

\begin{equation}
 p(t) = \begin{cases}
  1, & \text{if } t\bmod C < F,\\
  p_{\text{sparse}}, & \text{otherwise}.
 \end{cases}
\end{equation}

\subsection{Dynamic Token Gating}
\label{sec:token-gating}

For each sample $b$, we identify the indices of valid tokens excluding padding and special markers. Let $\mathcal{I}_b$ denote those indices and $N_b = |\mathcal{I}_b|$. We compute the combined difficulty $d_{b,i}$ for $i\in\mathcal{I}_b$ using Equation~\eqref{eq:combined} and select the top
$k_b = \max\bigl(1, \lfloor p(t)\cdot N_b\rfloor\bigr)$ tokens. The resulting binary mask $\boldsymbol{m}_{b}\in\{0,1\}^T$ is set to one for retained tokens and zero otherwise. We always retain all special visual tokens (e.g., those corresponding to a frame or image) regardless of their difficulty to preserve multimodal information.
Because the difficulty buffer $m_{s,i}$ is updated after every epoch, the set of selected positions adapts throughout training: tokens that become easy for the student are gradually deprioritized, while persistently challenging tokens or those requiring visual grounding remain active. This dynamic gating enables the model to allocate its computational budget to the most informative parts of the sequence at each epoch, rather than committing to a fixed sparsity pattern.
Finally, the per‑sample binary masks are concatenated and padded to form a batch mask $\mathbf{M}\in\{0,1\}^{B\times T'}$ where $T'$ is the expanded sequence length accounting for visual tokens.

\subsection{Adaptive Decoder Sparsity}

To exploit the binary mask during forward propagation, we modify the transformer decoder layer in the backbone LLM to support token-level sparse computation. Algorithm~\ref{algo:sparse} outlines how the binary mask $M$ governs computation within each decoder layer. 
Given the input hidden states, we first apply layer normalization, after which self-attention is computed only over tokens marked as active by $M$. In practice, sparse attention is implemented by directly passing $M$ as the attention mask to flash attention routines, zeroing out the hidden states of pruned tokens, and restricting query, key, and value computations to active positions only. 
Specifically, queries are formed only for retained tokens, while keys and values are gathered from the same active subset, excluding pruned tokens entirely from attention computation. The resulting attention outputs are then incorporated via residual connections.

We apply an analogous strategy to the feed-forward network. Hidden states corresponding to active tokens are gathered, processed by the MLP, and scattered back to their original positions. Tokens not selected by $M$ bypass the MLP and remain unchanged, with residual connections ensuring that their previous representations are propagated forward across layers. As a result, each decoder layer performs both self-attention and MLP only on the active subset, while inactive tokens are efficiently skipped without altering the model’s functional behavior. This design introduces no additional parameters, integrates seamlessly with existing frameworks such as HuggingFace Transformers, and remains fully compatible with pre-trained weights.

During back propagation, gradients are computed only for parameters associated with the active tokens. Inactive tokens are treated as constants, and their activations and losses receive no gradient updates. Residual connections preserve continuity of gradient flow across layers, aligning the backward computation with the sparse forward path and maintaining training stability in practice.

\begin{algorithm}[t]
\caption{Sparse Decoder Layer Forward}
\label{algo:sparse}
\begin{algorithmic}[1]
  \Require $\mathbf{H}\in\mathbb{R}^{B\times S\times D}$  \Comment{hidden states}
  \Require $\mathbf{M}\in\{0,1\}^{B\times S}$  \Comment{token mask}
  \For{$b = 1$ \textbf{to} $B$}              \Comment{$B$ = batch size}
    \State $\mathbf{x} \gets \mathrm{LN}_{\text{in}}(\mathbf{H}[b])$
    \State $\mathbf{mask} \gets \mathbf{M}[b]$ \Comment{$1$=keep, $0$=skip}
    \State $\mathbf{a} \gets \mathrm{SelfAttn}(\mathbf{x}, \mathbf{mask})$
    \State $\mathbf{H}[b] \gets \mathbf{H}[b] + \mathbf{a}$
    \State $\mathrm{active} \gets \mathrm{nonzero}(\mathbf{mask})$
    \State $\mathbf{h} \gets \mathrm{MLP}(\mathrm{LN}_{\text{post}}(\mathbf{H}[b])[\mathrm{active}])$
    \State $\mathbf{H}[b][\mathrm{active}] \gets
           \mathbf{H}[b][\mathrm{active}] + \mathbf{h}$
  \EndFor
  \State \Return $\mathbf{H}$
\end{algorithmic}
\end{algorithm}

\section{Experiments}

\subsection{Implementation Details}
\label{sec:implementation}



To demonstrate the effectiveness and generality of \textsc{ReGATE}, we apply it to three representative MLLMs, VideoChat2~\cite{li2024mvbench}, VideoLLaMA2~\cite{cheng2024videollama2}, and InternVL3.5~\cite{wang2025internvl35}, spanning diverse architectures, training paradigms, and scales. We exclude models such as Qwen2.5-VL~\cite{bai2025qwen25vl} and VideoLLaMA3~\cite{zhang2025videollama3} due to the lack of publicly available pretrained weights. Training such models from scratch is impractical, as they rely on web-scale data and hundreds of GPUs. However, with sufficient resources and pretrained weights, \textsc{ReGATE} can be seamlessly integrated into any MLLM training pipeline. For all experiments, the per-token reference losses from the text-only teacher are computed once over the entire fine-tuning dataset before student training begins and then cached and reused throughout training.


\noindent\textbf{VideoLLaMA2.} We first apply \textsc{ReGATE} to VideoLLaMA2-7B~\citep{cheng2024videollama2}, whose Qwen2-7B~\citep{yang2024qwen2} backbone is unfrozen during multimodal fine-tuning. \textsc{ReGATE} is introduced at this stage, using a text-only reference teacher (the same backbone without the visual encoder or adapter) that computes per-token losses with visual tokens masked out. This configuration allows us to examine how \textsc{ReGATE} interacts with full fine-tuning on a multimodal backbone.


\noindent\textbf{VideoChat2.}\ To evaluate \textsc{ReGATE} under parameter-efficient fine-tuning (PEFT), we integrate it into the LoRA-based Stage 3 training of VideoChat2-7B~\citep{li2024mvbench}, built on Mistral-7B~\citep{jiang2023mistral7b}. Gradients for LoRA parameters are computed only from the high-importance tokens selected by \textsc{ReGATE}, while the language backbone weights remain frozen. This setup enables us to assess whether \textsc{ReGATE} can complement PEFT methods without architectural modifications.


\noindent\textbf{InternVL3.5.}\ Finally, we evaluate scalability of \textsc{ReGATE} on InternVL3.5-14B~\citep{wang2025internvl35}, based on Qwen3-14B~\citep{yang2025qwen3technicalreport}. The reference teacher is derived from the same backbone by removing its visual encoder and projector, forming a text-only LLM. During fine-tuning, \textsc{ReGATE} dynamically gates tokens to focus computation on the most informative positions, reducing activation memory and training FLOPs. This setup allows us to assess \textsc{ReGATE}’s scalability and stability at large model scales.

\noindent\textbf{Datasets and sparsity schedule.} We fine-tune VideoLLaMA2 with and without \textsc{ReGATE} on the VideoChatGPT dataset~\citep{maaz2024videochatgpt}, which is a subset of VideoLLaMA2's official fine-tuning dataset containing approximately 300,000 instruction-response pairs. For VideoChat2, we similarly use a subset of its official fine-tuning data comprising around 2.6 million instruction pairs. For InternVL3.5, we adopt the same 2.6M-sample dataset used for VideoChat2, since the full InternVL3.5 training corpus is extremely large and not fully released. Training follows the dual-cycle sparsity schedule described in Section~\ref{sec:dual-cycle}, with parameters set to $C=128$, $F=16$, and $p_{\text{sparse}}=0.5$. To ensure stable training at the start, we prepend a global warm-up phase of 100 iterations, during which all tokens are retained. The main hyperparameters for \textsc{ReGATE} include an exponential moving average (EMA) decay of $\beta=0.9$ and a teacher loss weighting coefficient of $\lambda=0.5$. All experiments on VideoLLaMA2 and VideoChat2 are run on 4 H100 GPUs, while experiments on InternVL3.5 use 16 H100 GPUs, all under mixed-precision training.

\begin{table*}[t]
\centering
\caption{\textbf{Zero-shot evaluation results on image understanding benchmarks.} Previous best results are highlighted in \textbf{bold}, while \textsc{ReGATE}'s best results are  \underline{underlined}. $I$: SEED benchmark results are reported only for the image subset. For baseline models, scores are taken from their official publications where available.}
\small
\resizebox{\textwidth}{!}{
\renewcommand{\arraystretch}{1.1}
\begin{tabular}{l l l | c c c c c}
\toprule
\textbf{Model} & \textbf{LLM} & \textbf{Tokens} & \textbf{ScienceQA} & \textbf{MME} & \textbf{VizWiz} & \textbf{POPE} & $\textbf{SEED}^{I}$ \\ \midrule
\rowcolor{gray!15}\multicolumn{8}{c}{\textit{Open-source Models}} \\
\midrule
InstructBLIP~\citep{dai2023instructblip}    & Vicuna-7B  & --   & 60.5 & 254.3/1137.1 & 34.5 & 86.1 & 46.4 \\
LLaVA-1.5~\citep{liu2024llava}      & Vicuna-7B   & --   & 66.8 & 302.1/1506.2 & 50.0 & 85.9 & 66.1 \\
Qwen-VL-Chat~\citep{bai2023qwenvl}   & Qwen-7B  & --     & 68.2 & 392.1/1467.8 & 38.9 & 74.9 & 58.2 \\
LLaVA-1.6~\citep{liu2024llava}      & Vicuna-7B  & --    & 70.1 & --           & 57.6 & 86.5 & 70.2 \\
VILA1.5~\citep{lin2024vila}       & Llama-2-13B  & --     & 79.1 & 288.9/1429.3 & \textbf{60.6} & 84.2 & 62.8 \\
LLaVA-Next~\citep{liu2024llavanext} & Mistral-7B & --       & 73.0 & 308.9/1512.3 & --   & 87.3 & 72.4 \\
LLaVA-OneVision~\citep{llavaov} & Qwen2-7B & --    & \textbf{95.4} & 415.7/1577.8 & 53.0 & 87.4 & 75.4 \\
Qwen2.5-VL~\citep{bai2025qwen25vl}     & Qwen2.5-7B   & -- & 89.0 & 613.9/\textbf{1698.1} & --   & 85.9 & 77.0 \\ [0.25em]
\midrule
\rowcolor{gray!15}\multicolumn{8}{c}{\textit{Proprietary Models}} \\
\midrule
Claude3.7-Sonnet~\citep{anthropic2025claude37} & -- & --         & 90.9 & 649.6/1189.7 & --   & 82.4 & 74.3 \\
Gemini-1.5-Flash~\citep{geminiteam2024gemini15} & -- & --         & 83.3 & 488.6/1589.3 & --   & \textbf{88.5} & 75.0 \\
Gemini-1.5-Pro~\citep{geminiteam2024gemini15}     & -- & --       & 85.7 & 548.2/1562.4 & --   & 88.2 & 76.0 \\
GPT-4o~\citep{openai2024gpt4ocard}   & --  & --           & 90.1 & \textbf{719.3}/1609.4 & --   & 85.0 & 76.4 \\
GPT-4.1~\citep{openai2024gpt4ocard} & -- & --             & 92.8 & 673.9/1663.6 & --   & 86.4 & \textbf{78.0} \\ [0.25em]
\midrule
\rowcolor{gray!15}\multicolumn{8}{c}{\textit{Models w/wo \textsc{ReGATE}}} \\
\midrule
VideoChat2 & Mistral-7B & 3.93B & 40.8 & 314.6/1244.0 & 28.5 & 86.2 & 45.9 \\
\rowcolor{yellow!10}\textbf{VideoChat2-\textsc{ReGATE}} & Mistral-7B & 2.22B (\textcolor{darkgreen}{↓ 43.51\%})& 46.6$\scriptstyle\textcolor{darkgreen}{+5.8}$ & 360.7/1287.8$\scriptstyle\textcolor{darkgreen}{+46.1/+43.8}$ & 32.5$\scriptstyle\textcolor{darkgreen}{+4.0}$ & 85.1$\scriptstyle\textcolor{red}{-1.1}$ & 47.2$\scriptstyle\textcolor{darkgreen}{+1.3}$ \\
VideoLLaMA2 & Qwen2-7B & 83.82M & 61.4 & 376.4/1474.0 & 46.8 & 86.7 & 70.4  \\
\rowcolor{yellow!10}\textbf{VideoLLaMA2-\textsc{ReGATE}} & Qwen2-7B & 49.27M (\textcolor{darkgreen}{↓ 41.22\%})& 80.5$\scriptstyle\textcolor{darkgreen}{+19.1}$ & 391.1/1507.1$\scriptstyle\textcolor{darkgreen}{+14.7/+33.1}$ & 48.0$\scriptstyle\textcolor{darkgreen}{+1.2}$ & 87.5$\scriptstyle\textcolor{darkgreen}{+0.8}$ & 70.0$\scriptstyle\textcolor{red}{-0.3}$ \\

InternVL3.5 & Qwen3-14B & 3.96B & 93.3 & 681.6/1694.3 & 60.6 & 91.6 & 76.8 \\
\rowcolor{yellow!10}\textbf{InternVL3.5-\textsc{ReGATE}} & Qwen3-14B & 2.32B (\textcolor{darkgreen}{↓ 41.41\%})& \underline{94.4}$\scriptstyle\textcolor{darkgreen}{+1.1}$ & \underline{689.3}/\underline{1698.8}$\scriptstyle\textcolor{darkgreen}{+7.7/+4.5}$ & \underline{61.5}$\scriptstyle\textcolor{darkgreen}{+0.9}$ & \underline{93.1}$\scriptstyle\textcolor{darkgreen}{+1.5}$ & \underline{76.6}$\scriptstyle\textcolor{red}{-0.2}$ \\
\bottomrule
\end{tabular}}
\label{tab:image}
\end{table*}

\subsection{Benchmarks and Baselines}
\label{sec:benchamrks}

We evaluate \textsc{ReGATE} across image, long-video, and short-video understanding benchmarks under the LMMs-Eval~\cite{zhang2024lmmseval} framework.
For \textbf{image understanding}, we include ScienceQA~\citep{lu2022scienceqa}, MME~\citep{fu2024mme}, VizWiz~\citep{gurari2018vizwiz}, POPE~\citep{li2023pope}, and SEED~\citep{li2024seed}, covering scientific reasoning, perception, answerability, hallucination, and broad multimodal comprehension. For \textbf{long-video understanding}, we evaluate on Video-MME~\citep{fu2025videomme}, LongVideoBench~\citep{wu2024longvideobench}, MLVU~\citep{zhou2025mlvu}, and EgoSchema~\citep{mangalam2023egoschema}, all of which emphasize extended temporal reasoning, consistency, and grounding across minutes- to hour-long sequences. For \textbf{short-video understanding}, we adopt MVBench~\citep{li2024mvbench}, Perception Test~\citep{patraucean2023perceptiontest}, Vinoground~\citep{zhang2024vinoground}, and NExT-QA~\citep{xiao2021nextqa}, targeting event recognition, local temporal relations, counterfactual reasoning, and causal/temporal action reasoning in clips of tens of seconds.

We evaluate \textsc{ReGATE} against a diverse set of state-of-the-art models, including high-performing open-source families (LLaVA, Qwen) and proprietary models (Google Gemini, OpenAI GPT, Anthropic Claude). This selection spans multiple LLM backbones and sizes, ensuring robust comparisons across different tasks (Tables~\ref{tab:image}, \ref{tab:long-video}, \ref{tab:short-video}).

\subsection{Results}
\label{sec:experiment-result}

\begin{table*}[t]
\centering
\caption{\textbf{Zero-shot evaluation results on long video understanding benchmarks.} Previous best results are highlighted in \textbf{bold}, while \textsc{ReGATE}'s best results are  \underline{underlined}. $\dagger$ Results on Video-MME are reported without subtitles. For baseline models, scores are taken from their official publications when available.}
\small
\resizebox{\textwidth}{!}{
\renewcommand{\arraystretch}{1.1}
\begin{tabular}{l l l l | c c c c}
\toprule
\textbf{Model} & \textbf{LLM} & \textbf{Frames} & \textbf{Tokens} &  $\textbf{Video-MME}^{\dagger}$ & \textbf{LongVideoBench} & \textbf{MLVU} & \textbf{EgoSchema} \\ \midrule
\rowcolor{gray!15}\multicolumn{8}{c}{\textit{Open-source Models}} \\
\midrule

Video-LLaVA~\citep{lin2024videollava} & Vicuna-7B  & 8     & --     & 39.9 & 39.1 & 47.3 & 38.4    \\
LLaMA-VID~\citep{li2024llamavid}   & Llama-2-7B & 1fps & --      & 25.9 & --   & 33.2 & 38.5    \\
LLaVA-NeXT-Video~\citep{zhang2024llavanextvideo} & Vicuna-7B & 32 & --    & --   & 43.5 & --   & 43.9    \\ 
LLaVA-NeXT-Video~\citep{zhang2024llavanextvideo} & Qwen2-32B & 32 & --    & 60.2 & --   & 65.5 & 60.9  \\ 
VILA1.5~\citep{lin2024vila}     &  Llama-2-40B & 8 & --       & 60.1 & --   & 56.7 & 58.0  \\
LLaVA-OneVision~\citep{llavaov} &  Qwen2-7B & 32 & --     & 58.2 & 56.4 & 64.7 & 60.1  \\
Qwen2.5-VL~\citep{bai2025qwen25vl}  &    Qwen2.5-7B & -- & --     & 65.1 & 56.0 & 70.2 & 65.0   \\ 
VideoLLaMA3~\citep{zhang2025videollama3} &  Qwen2.5-7B  & 1fps  & --   & 66.2 & 59.8 & \textbf{73.0} & 63.3  \\ [0.25em]
\midrule
\rowcolor{gray!15}\multicolumn{8}{c}{\textit{Proprietary Models}} \\
\midrule
Gemini-1.5-Flash~\citep{geminiteam2024gemini15} & -- & -- & --         & 70.3  & 61.6   & --    & 65.7 \\
Gemini-1.5-Pro~\citep{geminiteam2024gemini15}   & -- & -- & --         & \textbf{75.0}  & 64.0   & --    & 71.2 \\
GPT-4o~\citep{openai2024gpt4ocard}      & -- & --  & --             & 71.9  & \textbf{66.7}   & 64.6  & \textbf{72.2} \\ [0.25em]
\midrule
\rowcolor{gray!15}\multicolumn{8}{c}{\textit{Models w/wo \textsc{ReGATE}}} \\
\midrule
VideoChat2 & Mistral-7B & 16 & 3.93B & 26.0 & 21.8 & 36.0 & 55.6  \\
\rowcolor{yellow!10}\textbf{VideoChat2-\textsc{ReGATE}} & Mistral-7B & 16 & 2.22B (\textcolor{darkgreen}{↓ 43.51\%})& 32.7$\scriptstyle\textcolor{darkgreen}{+6.7}$ & 24.3$\scriptstyle\textcolor{darkgreen}{+2.5}$ & 40.5$\scriptstyle\textcolor{darkgreen}{+4.5}$ & 54.8$\scriptstyle\textcolor{red}{-0.8}$ \\
VideoLLaMA2 & Qwen2-7B & 16 & 83.82M & 53.7 & 47.7 & 53.2 & 58.2  \\
\rowcolor{yellow!10}\textbf{VideoLLaMA2-\textsc{ReGATE}} & Qwen2-7B & 16 & 49.27M (\textcolor{darkgreen}{↓ 41.22\%})& 54.5$\scriptstyle\textcolor{darkgreen}{+0.8}$ & 47.6$\scriptstyle\textcolor{red}{-0.1}$ & 54.5$\scriptstyle\textcolor{darkgreen}{+1.3}$ & 56.4$\scriptstyle\textcolor{red}{-1.8}$ \\
InternVL3.5 & Qwen3-14B & 16 & 3.96B & 62.4 & 57.9 & 63.7 & 64.7 \\
\rowcolor{yellow!10}\textbf{InternVL3.5-\textsc{ReGATE}} & Qwen3-14B & 16 & 2.32B (\textcolor{darkgreen}{↓ 41.41\%})& \underline{63.0}$\scriptstyle\textcolor{darkgreen}{+0.6}$ & \underline{58.0}$\scriptstyle\textcolor{darkgreen}{+0.1}$ & \underline{64.2}$\scriptstyle\textcolor{darkgreen}{+0.5}$ & \underline{63.9}$\scriptstyle\textcolor{red}{-0.8}$ \\
\bottomrule
\end{tabular}}
\label{tab:long-video}
\end{table*}

\begin{figure}[t!]
    \centering
    \includegraphics[width=\linewidth]{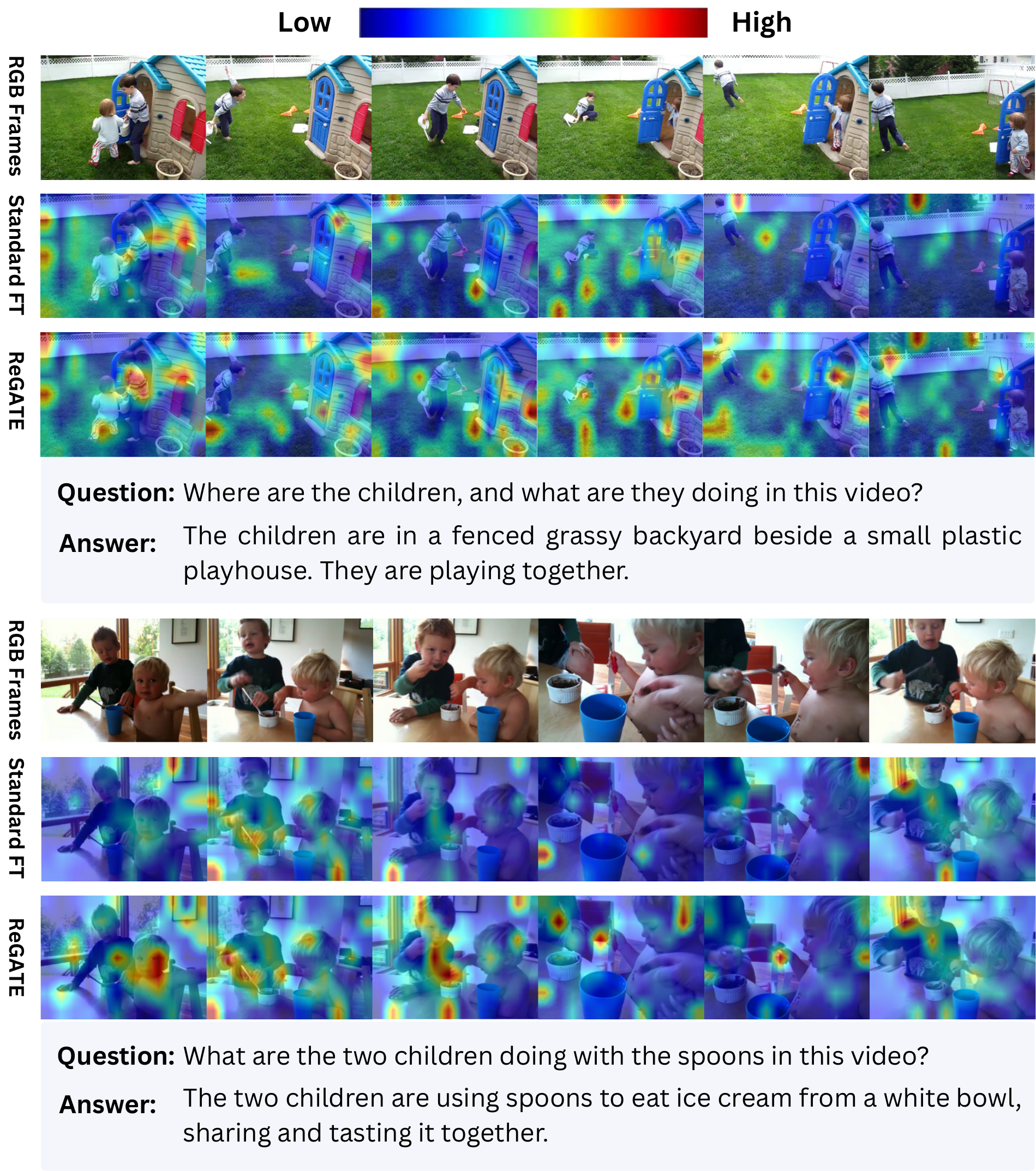}
    \caption{Attention maps from standard fine-tuning and \textsc{ReGATE} on video QA tasks. \textsc{ReGATE} focuses on contextually relevant regions (e.g., hands and manipulated objects), whereas standard fine-tuning spreads attention across the background.}
    \label{fig:saliency}
\end{figure}

\noindent\textbf{Learning better: ReGATE’s accuracy gains across image and video benchmarks.} The results presented in Tables~\ref{tab:image}, \ref{tab:long-video}, and \ref{tab:short-video} show how VideoLLaMA2, VideoChat2, and InternVL3.5 perform, with and without \textsc{ReGATE}, across image, short video, and long video understanding benchmarks. \textsc{ReGATE} improves performance consistently by focusing computation on the most informative tokens. Figure~\ref{fig:saliency} visualizes the learned attention maps, showing that ReGATE concentrates attention on visually and semantically important tokens compared to standard fine-tuning. For example, VideoLLaMA2-\textsc{ReGATE} outperforms the baseline VideoLLaMA2 on most tasks while using 41.22\% fewer tokens. Similarly, VideoChat2-\textsc{ReGATE} achieves better results than the baseline VideoChat2 while using 43.51\% fewer tokens.\ At the larger 14B scale, InternVL3.5-\textsc{ReGATE} also shows gains across both image and video benchmarks, confirming that \textsc{ReGATE} remains effective for high-capacity backbones.

On image understanding tasks requiring multimodal reasoning, all three models show significant gains.\ VideoLLaMA2-\textsc{ReGATE} improves by 19.1\% on ScienceQA and by up to 33.1 points on MME. VideoChat2-\textsc{ReGATE} improves by 5.8\% and 46.1 points on the same benchmarks. InternVL3.5-\textsc{ReGATE} also shows clear improvements, including +7.7/+4.5 points on MME, 0.9\% on VizWiz, and 1.6\% on POPE. For long-video understanding, VideoChat2-\textsc{ReGATE} shows strong improvements of 6.7\% on Video-MME and 4.5\% on MLVU. VideoLLaMA2-\textsc{ReGATE} also improves, though more modestly, with gains of 0.8\% and 1.3\% on the same tasks. At the larger scale, InternVL3.5-\textsc{ReGATE} further improves performance by 0.6\% on Video-MME, and 0.5\% on MLVU. Short video tasks benefit as well. VideoLLaMA2-\textsc{ReGATE} improves by 1.6\% on MVBench and 1.1\% on Perception, and VideoChat2-\textsc{ReGATE} gains 0.9\% and 1.6\%. InternVL3.5-\textsc{ReGATE} also shows improvements, with +1.3\% on MVBench and +1.4\% on Perception. 
Beyond performance metrics, \textsc{ReGATE} generalizes well across architectures and training strategies. It works effectively with different backbones such as Mistral, Qwen2, and Qwen3, and retains its benefits under both full-parameter and LoRA fine-tuning. These results confirm that \textsc{ReGATE}’s efficiency and accuracy gains are architecture- and training-strategy–agnostic, enabling seamless integration into diverse multimodal models.

\begin{table*}[t]
\centering
\caption{\textbf{Zero-shot evaluation results on short video understanding benchmarks.} Previous best results are highlighted in \textbf{bold}, while \textsc{ReGATE}'s best results are  \underline{underlined}. $\ddagger$ Results reported for Vinoground only for its video sub-task. For baseline models, scores are taken from their official publications when available.}
\small
\resizebox{\textwidth}{!}{
\renewcommand{\arraystretch}{1.1}
\begin{tabular}{l l l l | c c c c}
\toprule
\textbf{Model} & \textbf{LLM} & \textbf{Frames} & \textbf{Tokens} & \textbf{MVBench} & \textbf{Perception} & $\textbf{Vinoground}^{\ddagger}$ & \textbf{NExT-QA} \\ 
\midrule
\rowcolor{gray!15}\multicolumn{8}{c}{\textit{Open-source Models}} \\
\midrule
Video-LLaVA~\citep{lin2024videollava} & Vicuna-7B & 8     & --      & 41.0 & 44.3 & 25.8 & --    \\
LLaMA-VID~\citep{li2024llamavid}   & Llama-2-7B & 1fps & --      & 41.9 & 44.6 & --   & --    \\
LLaVA-NeXT-Video~\citep{zhang2024llavanextvideo} & Vicuna-7B & 32 & --    & 46.5 & 48.8 & 25.6 & --    \\ 
LLaVA-NeXT-Video~\citep{zhang2024llavanextvideo} & Qwen2-32B & 32 & --    & --   & 59.4 & --   & 77.3  \\ 
VILA1.5~\citep{lin2024vila}     &  Llama-2-40B & 8 & --       & --   & 54.0 & --   & 67.9  \\
LLaVA-OneVision~\citep{llavaov} &  Qwen2-7B & 32 & --     & 56.7 & 57.1 & 29.4 & 79.4  \\
Qwen2.5-VL~\citep{bai2025qwen25vl}  &    Qwen2.5-7B & -- & --     & 69.6 & 70.5 & --   & --    \\ 
VideoLLaMA3~\citep{zhang2025videollama3} &  Qwen2.5-7B  & 1fps  & --   & \textbf{69.7} & \textbf{72.8} & --   & \textbf{84.5}  \\ [0.25em]
\midrule
\rowcolor{gray!15}\multicolumn{8}{c}{\textit{Proprietary Models}} \\
\midrule
Gemini-1.5-Pro~\citep{geminiteam2024gemini15}   & -- & -- & --         & 60.5  & --   & 22.6  & -- \\
GPT-4o~\citep{openai2024gpt4ocard}      & -- & --  & --             & 64.6  & --   & \textbf{38.2}  & -- \\ [0.25em]
\midrule
\rowcolor{gray!15}\multicolumn{8}{c}{\textit{Models w/wo \textsc{ReGATE}}} \\
\midrule
VideoChat2 & Mistral-7B & 16 & 3.93B & 55.7 & 48.4 & 22.0 & 75.2  \\
\rowcolor{yellow!10}\textbf{VideoChat2-\textsc{ReGATE}} & Mistral-7B & 16 & 2.22B (\textcolor{darkgreen}{↓ 43.51\%})& 56.6$\scriptstyle\textcolor{darkgreen}{+0.9}$ & 50.0$\scriptstyle\textcolor{darkgreen}{+1.6}$ & 22.8$\scriptstyle\textcolor{darkgreen}{+0.8}$ & 75.5$\scriptstyle\textcolor{darkgreen}{+0.3}$ \\
VideoLLaMA2 & Qwen2-7B & 16 & 83.82M & 52.0 & 53.0 & 24.6 & 70.8  \\
\rowcolor{yellow!10}\textbf{VideoLLaMA2-\textsc{ReGATE}} & Qwen2-7B & 16 & 49.27M (\textcolor{darkgreen}{↓ 41.22\%})& 53.6$\scriptstyle\textcolor{darkgreen}{+1.6}$ & 54.1$\scriptstyle\textcolor{darkgreen}{+1.1}$ & 25.2$\scriptstyle\textcolor{darkgreen}{+0.6}$ & 70.0$\scriptstyle\textcolor{red}{-0.8}$ \\
InternVL3.5 & Qwen3-14B & 16 & 3.96B & 68.3 & 65.3 & 31.2 & 80.8 \\
\rowcolor{yellow!10}\textbf{InternVL3.5-\textsc{ReGATE}} & Qwen3-14B & 16 & 2.32B (\textcolor{darkgreen}{↓ 41.41\%})& \underline{69.6}$\scriptstyle\textcolor{darkgreen}{+1.3}$ & \underline{66.7}$\scriptstyle\textcolor{darkgreen}{+1.4}$ & \underline{31.2}$\scriptstyle\textcolor{darkgreen}{+0.0}$ & \underline{81.2}$\scriptstyle\textcolor{darkgreen}{+0.4}$ \\
\bottomrule
\end{tabular}}
\label{tab:short-video}
\end{table*}

\begin{table*}[t]
  \centering
  \small
  \vspace{0.3cm}
  \caption{\textbf{Efficiency comparison of different models with \textsc{ReGATE}.} 
  Performance is measured as the average zero-shot accuracy (\%) on video benchmarks. We report one-time teacher overhead and training time as GPU-hours (wall-clock hours $\times$ \#GPUs).}
  \label{tab:efficiency}
  \resizebox{\textwidth}{!}{
  \begin{tabular}{l c c c c c}
    \toprule
    \textbf{Model} & \textbf{Tokens $\downarrow$} & \textbf{Teacher Cost (GPU-h)$\downarrow$} & \textbf{Train Time (GPU-h)$\downarrow$} & \textbf{Avg. Mem/GPU (GB)$\downarrow$} & \textbf{Avg. Acc. (\%) $\uparrow$} \\
    \midrule
    VideoLLaMA2 & 83.82M & - & 129.6 & 69.1 & 48.2 \\
    \rowcolor{blue!10}\textbf{VideoLLaMA2-\textsc{ReGATE}} & 49.27M & 2.1 & 107.6 & \textbf{61.3} & \textbf{48.9} \\
    \rowcolor{yellow!10}\textbf{VideoLLaMA2-\textsc{ReGATE}} & \textbf{29.32M} & 2.1 & \textbf{64.0} & - & 48.0 \\
    \midrule
    VideoChat2 & 3.93B & - & 148.8 & 70.8 & 46.1 \\
    \rowcolor{blue!10}\textbf{VideoChat2-\textsc{ReGATE}} & 2.22B  & 10.0 & 130.0 & \textbf{63.7} & \textbf{47.8} \\
    \rowcolor{yellow!10}\textbf{VideoChat2-\textsc{ReGATE}} & \textbf{1.51B} & 10.0 & \textbf{86.4} & - & 46.0 \\
    \midrule
    InternVL3.5 & 3.96B & - & 435.2 & 58.3 & 61.8 \\
    \rowcolor{blue!10}\textbf{InternVL3.5-\textsc{ReGATE}} & 2.32B  & 11.3 & 374.4 & \textbf{51.9} & \textbf{62.2} \\
    \rowcolor{yellow!10}\textbf{InternVL3.5-\textsc{ReGATE}} & \textbf{1.63B} & 11.3 & \textbf{262.4} & - & 61.6 \\
    \bottomrule
  \end{tabular}}
\end{table*}

\noindent\textbf{Learning faster: ReGATE’s efficiency gains.} 
Table~\ref{tab:efficiency} summarizes gains in token usage, training time, and accuracy on video benchmarks. Since \textsc{ReGATE} introduces no architectural changes, per-layer FLOPs remain identical to the baseline.

For VideoLLaMA2-7B, \textsc{ReGATE} matches baseline accuracy (48.0\% vs.\ 48.2\%) in only 64.0 GPU-hours, less than half the standard fine-tuning time (129.6 GPU-hours), while using 29.32M tokens (35\% of the baseline’s 83.82M). Extending training to 107.6 GPU-hours, still 22 hours fewer than the baseline, it processes 41.5\% fewer tokens and surpasses the baseline with 48.9\% accuracy. The one-time teacher pass adds only 2.1 GPU-hours, less than 2\% of the baseline cost.

For VideoChat2-7B, which employs LoRA fine-tuning, time savings are smaller due to its already efficient backward pass. \textsc{ReGATE} matches baseline accuracy (46.0\% vs.\ 46.1\%) in 86.4 GPU-hours, compared to 148.8 for the baseline, using 38\% of the baseline tokens (1.51B vs.\ 3.93B). When extended to 130.0 GPU-hours, still 18.8 fewer than baseline, it processes 43.5\% fewer tokens (2.22B vs.\ 3.93B) and improves accuracy to 47.8\%. The teacher pass costs only 10.0 GPU-hours, about 7\% of baseline.

For the larger InternVL3.5-14B, \textsc{ReGATE} maintains consistent gains. With aggressive pruning, it achieves near-baseline accuracy (61.6\% vs.\ 61.8\%) in 262.4 GPU-hours, a 40\% speed-up over the baseline, using just 41\% of baseline tokens (1.63B vs.\ 3.96B). Extending to 374.4 GPU-hours, it reduces token usage by 41.4\% (2.32B vs.\ 3.96B) and surpasses the baseline at 62.2\% accuracy. The teacher pass requires only 11.3 GPU-hours, less than 3\% of the baseline cost.

This speed-up difference across models stems from variations in training strategies and model scales. In full fine-tuning, as used in VideoLLaMA2, both forward and backward passes are computationally expensive. By pruning tokens, \textsc{ReGATE} accelerates both passes, particularly the backward pass where gradients are computed for all parameters. In LoRA fine-tuning, as in VideoChat2, most parameters are frozen, so the backward pass is already efficient; \textsc{ReGATE} primarily speeds up the forward pass, resulting in smaller total time savings.
At the 14B scale, as in InternVL3.5, additional bottlenecks such as GPU memory and inter-GPU communication arise. \textsc{ReGATE} reduces both the number of processed tokens and the size of intermediate activations, alleviating memory and communication costs. Consistent with these reductions, it lowers average per-GPU memory by 7--8 GB (about 10–11\%) across all backbones: $69.1\rightarrow61.3$ GB on VideoLLaMA2, $70.8\rightarrow63.7$ GB on VideoChat2, and $58.3\rightarrow51.9$ GB on InternVL3.5. Overall, \textsc{ReGATE} delivers significant efficiency gains across fine-tuning regimes and model scales, providing a flexible and effective approach for accelerating multimodal training without compromising performance.


\begin{figure*}[h!]
    \centering
    \includegraphics[width=\linewidth]{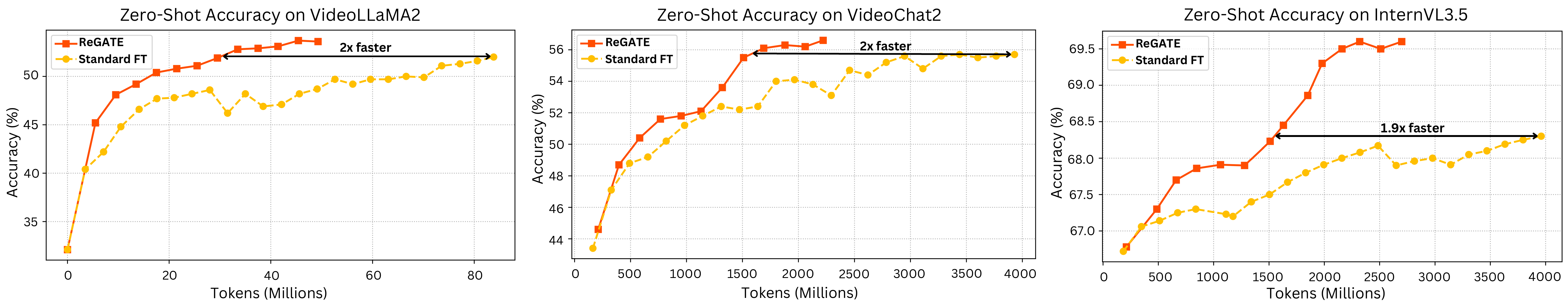}
    \caption{\textbf{Zero-shot accuracy on MVBench during fine-tuning.} 
    \textsc{ReGATE} (red) consistently outperforms standard fine-tuning (orange) at the same token count. It reaches the baseline’s peak accuracy roughly twice as fast while using only 38\% of the tokens on average, and surpasses the baseline with 41\% fewer tokens.}
    \label{fig:intro_mvbench}
\end{figure*}

\noindent\textbf{How does \textsc{ReGATE} compare with state-of-the-art training-time efficiency methods?}
We compare \textsc{ReGATE} with two representative approaches: LaVi~\citep{yue2025lavi} for video understanding and LLaVA-Meteor~\citep{li2025llavameteror} for image understanding. LaVi introduces additional parameters and architectural changes and must be trained from scratch on the backbone, whereas \textsc{ReGATE} is parameter-free and architecture-agnostic. For images, LLaVA-Meteor prunes tokens using heuristics and a tuned hyperparameter, while \textsc{ReGATE} is fully self-adaptive, guided by token-level learning difficulty. All comparisons use comparable backbone sizes to ensure fairness.
On video benchmarks (Table~\ref{tab:lavi}), despite using the weaker VideoLLaMA2-7B backbone, \textsc{ReGATE} achieves competitive or superior results on two of four datasets compared to LaVi. On image benchmarks (Table~\ref{tab:meteor}), when applied to InternVL3.5-14B, \textsc{ReGATE} consistently outperforms Meteor. These results highlight that \textsc{ReGATE} delivers strong training-time efficiency without architectural changes or parameter overhead, while maintaining or improving accuracy across both video and image understanding tasks.
\begin{table}[h!]
\centering
\caption{\textbf{Comparison with LaVi.} LaVi results are reported from the original paper.}
\resizebox{0.475\textwidth}{!}{%
\begin{tabular}{lccccc}
\toprule
Method & LLM & VideoMME & MLVU & EgoSchema & MVBench \\
\midrule
LLaVA-OneVision + LaVi & Qwen2-7B & 54.0 & \textbf{58.5} & 55.5 & \textbf{54.3} \\
\rowcolor{yellow!10}VideoLLaMA2 + \textsc{ReGATE} & Qwen2-7B & \textbf{54.5} & 54.5 & \textbf{56.4} & 53.6 \\
\bottomrule
\end{tabular}}
\label{tab:lavi}
\end{table}

\begin{table}[h!]
\centering
\caption{\textbf{Comparison with LLaVA-Meteor.} LLaVA-Meteor results are from the original paper.}
\resizebox{0.475\textwidth}{!}{%
\begin{tabular}{lcccc}
\toprule
Method & LLM & VizWiz & POPE & SEED \\
\midrule
LLaVA-Meteor & Vicuna-13B & 55.3 & 87.2 & 64.8 \\
\rowcolor{yellow!10}InternVL3.5 + \textsc{ReGATE} & Qwen3-14B & \textbf{60.5} & \textbf{93.1} & \textbf{76.6} \\
\bottomrule
\end{tabular}}
\label{tab:meteor}
\end{table}

\section{Conclusion}

We introduced \textsc{ReGATE}, a reference-guided token gating framework that accelerates the training of multimodal large language models. By combining a student model's learning difficulty with reference losses from a frozen text-only teacher, \textsc{ReGATE} dynamically allocates computation to the most informative tokens while skipping those less relevant for multimodal understanding. The method is simple to implement, requires no architectural changes, and substantially improves training efficiency. Experiments show that \textsc{ReGATE} achieves comparable or better accuracy than standard full fine-tuning, using only a fraction of the tokens and significantly less training time. These results suggest that \textsc{ReGATE} provides a practical and general solution for improving MLLM training efficiency. Future work will explore adaptive scheduling for token sparsity by dynamically adjusting the retained token ratio based on task complexity, model stability, and training progress, starting with higher sparsity early and gradually relaxing it during fine-tuning.

\newpage

\section*{Limitations}

A limitation of the current \textsc{ReGATE} is that it uses a fixed design for both sparsity scheduling and reference supervision. While this choice keeps the training pipeline simple and stable, it may leave room for improvement in more challenging settings. For example, dynamically adjusting the retained token ratio across training or according to input complexity could yield a better efficiency–performance trade-off. Similarly, extending the reference from a frozen text-only teacher to stronger teachers may provide richer guidance, particularly for tasks requiring fine-grained spatial or temporal reasoning.

\section*{Acknowledgments}
This research was supported by the National Eye Institute (NEI) of the National Institutes of Health (NIH) under award number R01EY034562.\ The content is solely the responsibility of the authors and does not necessarily represent the official views of the NIH. 

\balance
\bibliography{custom}

\clearpage
\appendix

\section*{Appendix}
\label{sec:appendix}

\section{Additional Ablation Studies}
\label{sec:ablation}

In this section, we delve deeper into the design choices of \textsc{ReGATE}, specifically examining the sensitivity of the weighting hyperparameter and the impact of the reference teacher's capacity.

\subsection{Impact of the Weighting Factor \texorpdfstring{$\lambda$}{lambda}}

\begin{table}[h]
\centering
\caption{\textbf{Ablation study on the weighting factor $\lambda$.} This parameter balances the student’s EMA-based difficulty and the teacher’s reference loss. Performance is reported as average zero-shot accuracy (\%) on eight video benchmarks.}
\label{tab:ablation_lambda}
\begin{tabular}{l | l | c}
\toprule
$\lambda$ & \textbf{Description} & \textbf{Acc. (\%)} \\ \midrule
$\lambda = 0.0$ & Student EMA Only & 47.7 \\
$\lambda = 1.0$ & Reference Loss Only & 46.4 \\
$\lambda = 0.5$ & \textbf{Combined Signals} & \textbf{48.9} \\ \bottomrule
\end{tabular}
\end{table}

To validate the contributions of the individual components within our dual-signal token scoring mechanism, we conduct an ablation study on the hyperparameter $\lambda$. This coefficient balances the two core signals in our difficulty score formulation: $d_{b,i} = m_{s,i} + \lambda \, \ell_{b,i}^{\text{ref}}$, where $m_{s,i}$ is the student's dynamic EMA difficulty and $\ell_{b,i}^{\text{ref}}$ is the static reference loss from the teacher model. We evaluate three values for $\lambda$: 0.0, 0.5, and 1.0. The experiments use our VideoLLaMA2-\textsc{ReGATE} setup with all other hyperparameters fixed for a fair comparison.

As shown in Table~\ref{tab:ablation_lambda}, setting $\lambda = 0.5$ achieves the best balance. Relying solely on the student's historical difficulty ($\lambda=0.0$) fails to capture zero-shot visual dependencies, while relying exclusively on the teacher ($\lambda=1.0$) ignores the student's training dynamics. The combined signal effectively isolates tokens that are both visually critical and challenging for the current model state.

\subsection{Impact of Reference Teacher Capacity}
\label{sec:teacher_capacity}

A critical design choice in \textsc{ReGATE} is the selection of the reference teacher. We investigate how the teacher model's capacity affects the student's performance. Specifically, for the VideoLLaMA2-7B~\cite{cheng2024videollama2} student (based on Qwen2-7B~\cite{yang2024qwen2}), we evaluate reference signals from a smaller model (Qwen2-1.5B) and a substantially larger model (Qwen2-57B-A14B), comparing them against the matched Qwen2-7B baseline. To ensure fair comparison, we restrict the study to a single model family, which guarantees consistent tokenization and avoids misalignment between teacher and student. Cross-architecture teachers (e.g., Mistral guiding Qwen) are excluded because distinct tokenizers yield inconsistent token boundaries for the same text, violating the one-to-one mapping required for per-token reference loss computation in \textsc{ReGATE}. Such a mismatch renders the reference signal mathematically invalid. 

\begin{table}[h]
    \centering
    \setlength{\tabcolsep}{3.5pt}
    \caption{\textbf{Impact of Reference Teacher Capacity.} We compare teachers of varying sizes for a VideoLLaMA2-7B student. Performance is reported as average zero-shot accuracy (\%) on eight video benchmarks.}
    \begin{tabular}{l|c}
    \toprule
    \textbf{Teacher Model} & \textbf{Avg. Acc. (\%)} \\
    \midrule
    Qwen2-1.5B & 45.4 \\
    Qwen2-57B & 46.8 \\
    Qwen2-7B (\textbf{Ours}) & \textbf{48.9} \\
    \bottomrule
    \end{tabular}
    \label{tab:teacher_size}
\end{table}

The results in Table~\ref{tab:teacher_size} show that the matched Qwen2-7B teacher outperforms both the smaller 1.5B and the much larger 57B models. This pattern reveals a capacity mismatch effect: performance degrades when the teacher’s capability diverges too far from the student’s. When using the large Qwen2-57B teacher, performance drops because the teacher is too strong relative to the student. With extensive world knowledge, the 57B model can often predict complex tokens from text alone, resulting in very low reference loss. However, the 7B student lacks this prior knowledge and depends on visual grounding to infer these tokens. As a result, the teacher mistakenly signals \textsc{ReGATE} to prune these positions, causing the student to skip crucial computations. Conversely, the smaller Qwen2-1.5B teacher underestimates the student’s ability. It assigns high loss to linguistic patterns or factual details that the 7B student already handles easily. This floods the top-$k$ selection with linguistically difficult but visually irrelevant tokens, wasting computation on trivial cases rather than focusing on visual reasoning. 

Overall, these findings suggest that the optimal reference teacher should be capacity-aligned with the student model to provide a meaningful and reliable learning signal. The reference loss is most effective when it closely reflects the student’s uncertainty, directing computation toward the tokens where textual priors alone fall short.

\section{Additional Benchmarks Details}
\label{app:benchmark_details}

Table~\ref{tab:bench-summary} lists the evaluation prompts for each benchmark used in the experiments, most of which are adapted from LMMs-Eval~\cite{zhang2024lmmseval}.

\newlength{\myresponsecolwidth}
\settowidth{\myresponsecolwidth}{Answer with the option’s letter from the given choice2123131231123123}

\begin{table*}[t]
  \centering
  \small
  \renewcommand{\arraystretch}{1.1} 
  \caption{Prompts specifying the response format used for each evaluation benchmark.}
  \label{tab:bench-summary}
  \begin{tabular}{@{} l | m{\myresponsecolwidth} @{}}
    \toprule
    \textbf{Benchmark} & \textbf{Response formatting prompts} \\
    \midrule
    POPE & -- \\[2pt]
    MME & Answer the question using a single word or phrase. \\[2pt]
    VisWiz & Answer the question using a single word or phrase. 
       When the provided information is insufficient, respond with ``Unanswerable''. \\[2pt]
    ScienceQA & Answer with the option’s letter from the given choices directly. \\[2pt]
    SEED & Answer with the option’s letter from the given choices directly. \\ [2pt]
    \midrule
    MLVU & -- \\[2pt]
    MVBench & Only give the best option. \\[2pt]
    Video-MME & Answer with the option’s letter from the given choices directly. \\[2pt]
    EgoSchema & Answer with the option’s letter from the given choices directly. \\[2pt]
    NExT-QA & -- \\ [2pt]
    Perception & Answer with the option's letter from the given choices directly. \\ [2pt]
    Vinoground & Please only output one English character. \\ [2pt]
    LongVideoBench & Answer with the option's letter from the given choices directly. \\ [2pt]
    \bottomrule
  \end{tabular}
\end{table*}

\section{Qualitative Analysis of Reference Loss}
\label{sec:qualitative}

\begin{figure*}[h!]
    \centering
    \includegraphics[width=1.0\linewidth]{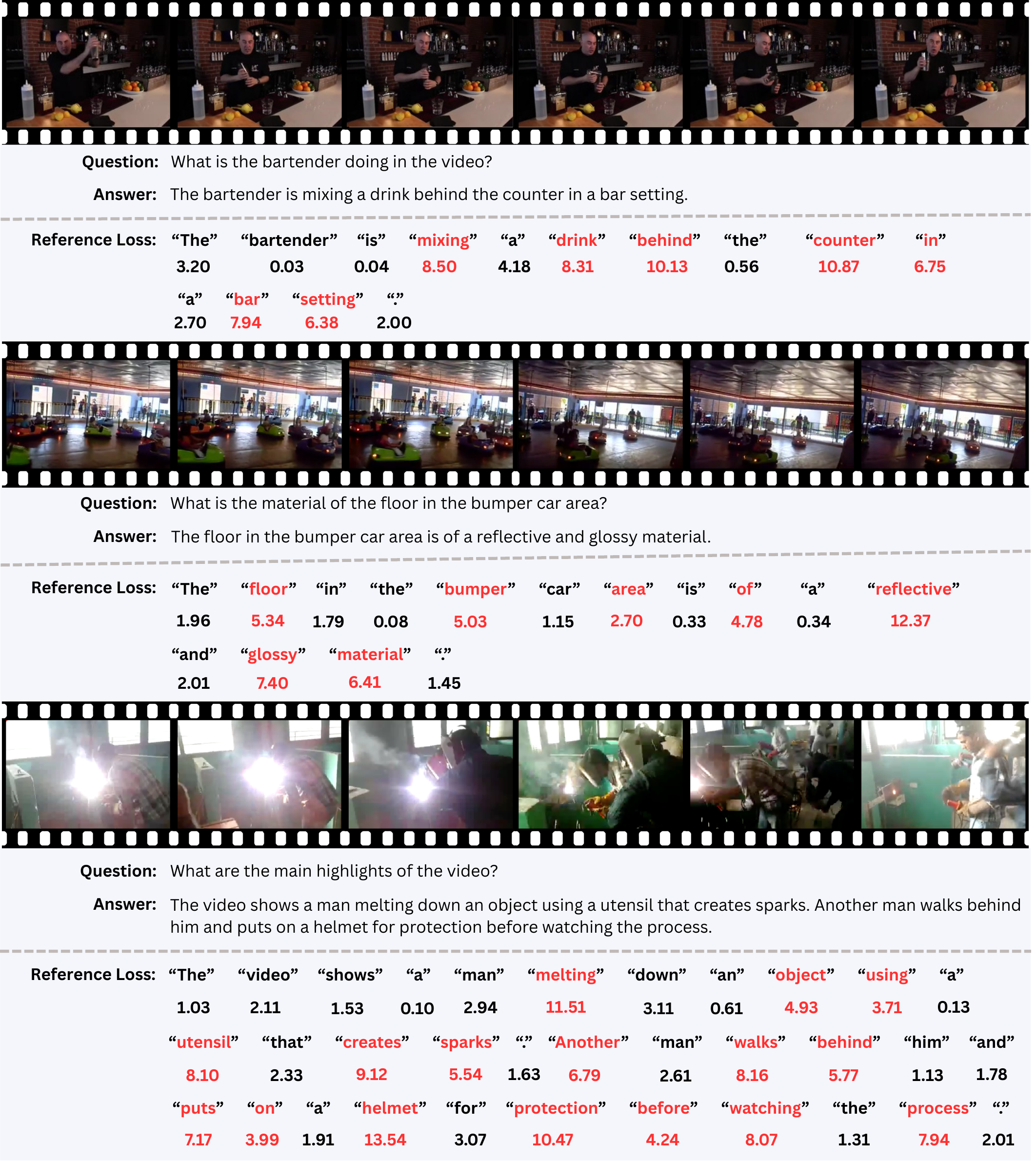}
    \caption{\textbf{Qualitative examples illustrating the effectiveness of the reference loss signal.} For two video Q\&A pairs, we show the per-token reference loss computed by a text-only teacher model (Mistral-7B). Tokens colored in \textbf{\textcolor{red}{red}} have the highest losses and represent the top 50\% most difficult tokens to predict from text alone. These are precisely the tokens that \textsc{ReGATE} prioritizes for computation.}
    \label{fig:qualitative}
\end{figure*}

To validate the core mechanism of \textsc{ReGATE}, we qualitatively analyze the reference loss signal that guides token selection. We assume that a high loss score from the text-only teacher indicates that a token requires visual information to be understood. Figure~\ref{fig:qualitative} shows two video Q\&A examples, visualizing the loss for each word in the answer as using a Mistral-7B~\citep{jiang2023mistral7b} teacher model.

The results strongly support our assumption. As illustrated in the figure, tokens for visual details that are hard to guess from text alone, like the action ``mixing'' or the attribute ``reflective'', get high loss scores. In contrast, simple grammatical words like ``The'' and ``is'', or terms repeated from the question like ``bartender'', get low scores. This difference confirms that reference loss is a reliable indicator of visual importance, enabling \textsc{ReGATE} to focus its computation on the most critical tokens for more efficient training.

\clearpage

\begin{figure*}[h!]
    \centering
    \includegraphics[width=1.0\linewidth]{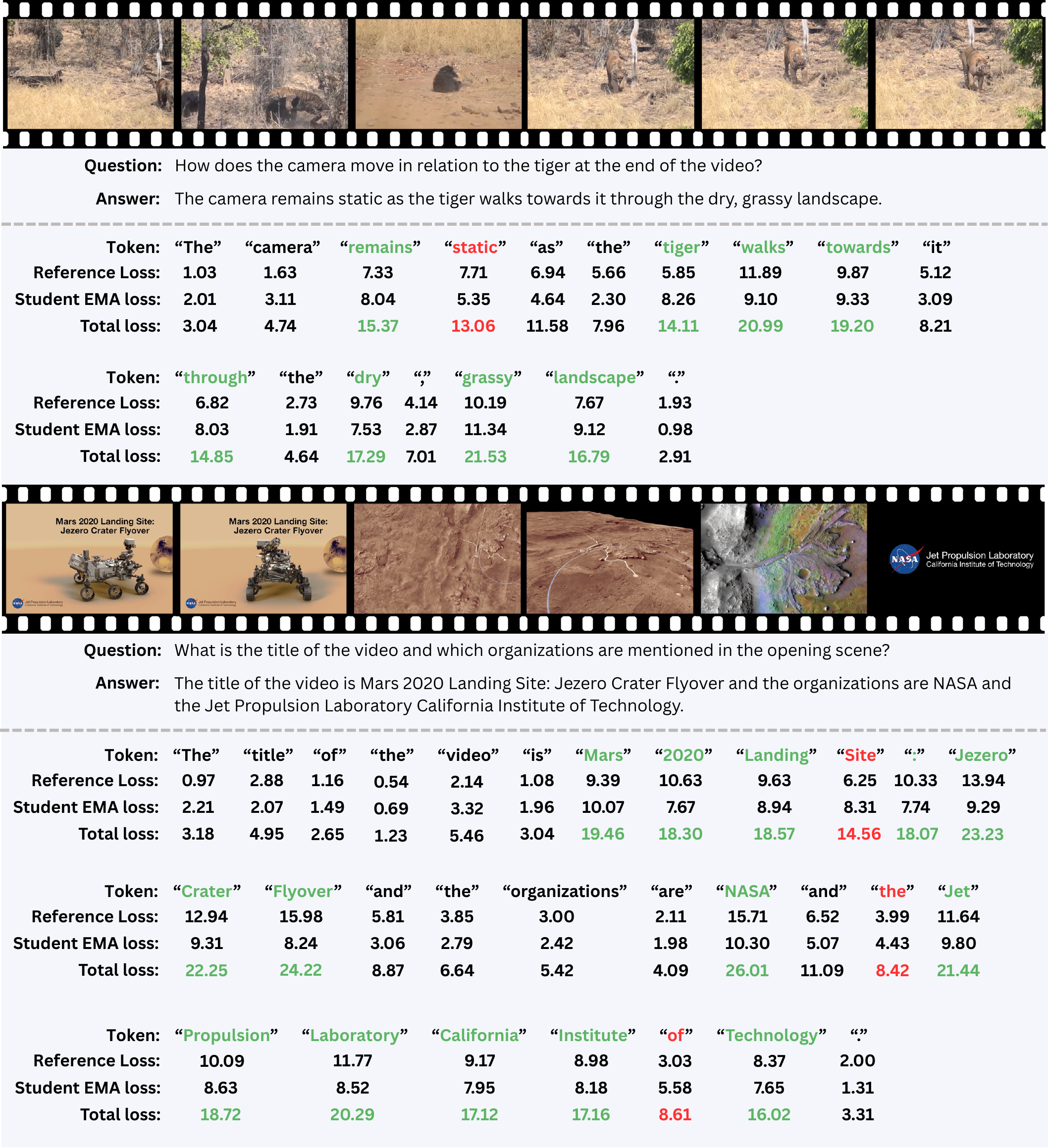}
    \caption{\textbf{Failure cases showing the limitation of fixed sparsity.} Tokens in \textbf{\textcolor{OliveGreen}{green}} are retained by \textsc{ReGATE}, while those in \textbf{\textcolor{red}{red}} are clearly helpful for answering but were not selected due to the fixed sparsity ratio ($p{=}0.5$). Some visually or semantically important tokens (e.g., static, NASA) are skipped, revealing the need for adaptive sparsity.}
    \label{fig:failure}
\end{figure*}

\end{document}